\documentclass[twoside]{article}
\usepackage[preprint]{aistats2026}
\usepackage{graphicx}
\usepackage{booktabs}
\usepackage{multirow}
\usepackage{hyperref}
\usepackage{url}
\usepackage{amsfonts}
\usepackage{subcaption}
\usepackage{xcolor}
\usepackage{bbm}

\usepackage{amsmath,amsfonts, amsthm}

\begin{document}



\twocolumn[

\aistatstitle{There is  More to Attention: Statistical Filtering Enhances Explanations in Vision Transformers}

\aistatsauthor{ Meghna P Ayyar \And Jenny Benois-Pineau \And  Akka Zemmari }

\aistatsaddress{ LaBRI, CNRS, Univ. Bordeaux, UMR 5800, F-33400, Talence, France} ]

\begin{abstract}
Explainable AI (XAI) has become increasingly important with the rise of large transformer models, yet many explanation methods designed for CNNs transfer poorly to Vision Transformers (ViTs). Existing ViT explanations often rely on attention weights, which tend to yield noisy maps as they capture token-to-token interactions within each layer.While attribution methods incorporating MLP blocks have been proposed, we argue that attention remains a valuable and interpretable signal when properly filtered. We propose a method that combines attention maps with a statistical filtering, initially proposed for CNNs, to remove noisy or uninformative patterns and produce more faithful explanations. We further extend our approach with a class-specific variant that yields discriminative explanations. Evaluation against popular state-of-the-art methods demonstrates that our approach produces sharper and more interpretable maps. In addition to perturbation-based faithfulness metrics, we incorporate human gaze data to assess alignment with human perception, arguing that human interpretability remains essential for XAI. Across multiple datasets, our approach consistently outperforms or is comparable to the SOTA methods while remaining efficient and human plausible.

\end{abstract}

\begin{figure*}[t]
    \centering
    \includegraphics[width=\linewidth]{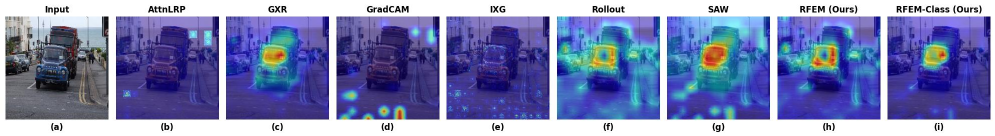}
    \caption{Explanation maps of (a) SOTA AttnLRP (b) Gradient $\times$ Relevance, (d) GradCAM, (e) Input $\times$ Gradient, (f)Attention Rollout, (g) SAW and our methods (h) RFEM and the class-specific (i) RFEM-Class}
    \label{fig:XAI_maps_sample}
\end{figure*}

\section{Introduction}\label{sec:intro}

Deep neural networks (DNNs) continue to achieve state-of-the-art performance across domains by encoding complex patterns into high-dimensional representations, but their success comes at the cost of interpretability. Without explicit mechanisms to reveal how decisions are made, DNNs are often perceived as opaque “black boxes”~\cite{castelvecchi2016can}. In safety-critical applications, such as healthcare~\cite{ali2023enlightening}, autonomous driving~\cite{kuznietsov2024explainable} etc., users require not only accurate predictions but also the reasoning behind them. Explanations of a model are therefore essential both for fostering trust~\cite{ferrario2022explainability} and for ensuring that models behave reliably and fairly under real-world conditions.

Early XAI research for computer vision focused on convolutional neural networks (CNNs), where the methods measured the contributions of each pixel in the input image to the output decision~\cite{selvaraju2017grad,chattopadhay2018grad,binder2016layer}. With the advent of Vision Transformers (ViTs)~\cite{dosovitskiy2020image}, traditional CNN-based XAI methods have faced challenges in providing high-quality explanations. A core component of Transformers is the self-attention mechanism, which enables the model to assess and weigh the importance of each input token relative to all other tokens. In ViTs, these attention values are computed across patches of an image, forming attention maps that reflect how much each patch ``attends to" others during feature extraction. Methods such as Attention Rollout\cite{abnar2020quantifying} assume that raw attention weights can serve as explanations. They aggregate attention matrices across layers to produce global explanation maps that reflect the model’s overall focus. Layer-wise attention maps are also widely used for visualising transformer behaviour, often focusing on individual layers or heads~\cite{attntransformer2019}. Another option is the aggregation by averaging of maps or multiplying them across layers to get a global view, as in attention rollout~\cite{abnar2020quantifying}. However, this naive averaging can blur important patterns, overlook the semantic roles of attention across different layers and heads, and lack class specificity. 

These shortcomings fuel an ongoing debate on whether attention constitutes a valid explanation. Critics argue that attention weights are not faithful, since shuffling them often does not affect model predictions~\cite{jainwallace2019}. Broader analyses~\cite{lopardo24posthoc} show that attention-based and post-hoc methods (e.g., gradient- or perturbation-based) can produce related but distinct explanations, often revealing complementary insights. However, ~\cite{bibal2022attention} argue that while attention may not constitute \textit{the} explanation, it can serve as \textit{an} explanation of model behaviour. This distinction aligns with the broader XAI discourse, where different paradigms prioritise different criteria for explanation quality. Much of the literature emphasises faithfulness, i.e., whether masking important regions changes model outputs~\cite{jainwallace2019}. Yet, faithfulness alone is insufficient. Plausibility, i.e. the degree to which explanations align with human intuition, is equally important~\cite{wiegreffepinter2019attention}. From this perspective, explanations that are not strictly faithful may still provide valuable insights to end users. 

Motivated by the dual goals of faithfulness and plausibility, we revisit attention as a basis for explanations in ViTs. Rather than discarding attention as an explanation, we propose to enhance it through statistical filtering. Our approach builds on the Feature Explanation Method (FEM)~\cite{FuadMGBBZ20}, which has been applied to medical imaging tasks~\cite{ayyar2023feature} for CNNs. In addition to standard faithfulness evaluation, we also assess plausibility using datasets with human gaze fixation data as in ~\cite{BourrouxBBG22}, treating gaze as a proxy for human interpretability.

Our main contributions are:
\begin{itemize}
    \item We introduce a statistical filtering approach for combining attention maps across Transformer layers, yielding explanation maps that are both faithful and human-plausible.
    \item We extend this approach with a class-specific variant to generate object-focused explanations.
    \item We evaluate our methods against five representative ViT XAI approaches using two datasets with gaze fixation ground truth for plausibility, alongside standard faithfulness and stability metrics. 
\end{itemize}

An illustrative comparison of the XAI maps for our methods is shown in Figure~\ref{fig:XAI_maps_sample} (h) and (i). The remainder of the paper is organised as follows. Overview of ViT architecture and related work are presented in Section~\ref{sec:related_work}, our method and its class-dependent variant in Section~\ref{sec:methodology}, experiments and results with details about the datasets and metrics used are reported in Section~\ref{sec:exp_results} and the conclusion and outline for future perspectives are given in Section~\ref{sec:conclusion}. 

\begin{figure*}
\begin{center}
    \includegraphics[width=0.9\linewidth]{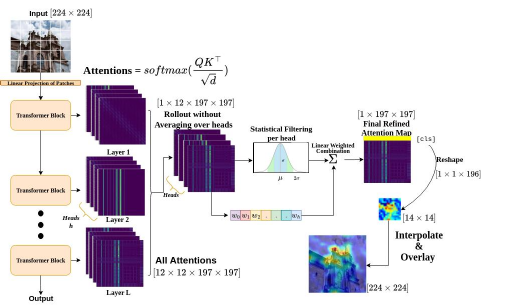} 
\caption{Pipeline for our RFEM method. Dimensions correspond to a ViT-B16 model with 12 layers and 12 heads per layer. The yellow row highlights the \textit{query} associated with the \texttt{[CLS]} token, used to generate the final explanation} \label{fig:rfem_framework}
\end{center}
\end{figure*}

\section{Related Work}\label{sec:related_work}

\textbf{Background} : Vision Transformers (ViTs) process an image by dividing it into $N \times N$ patches, embedding each patch as a token, and adding a learnable \texttt{[CLS]} token to summarise the image. The model consists of $L$ transformer layers, each with $H$ attention heads, which compute pairwise interactions among all tokens. Within each layer, tokens are projected into queries (Q), keys (K), and values (V), and the output of the Multi-Head Attention (MHA) is computed as $\text{softmax}(QK^\top/\sqrt{d})V$, where $d$ denotes the dimensions of the token embeddings. This is followed by a residual connection, Layer Normalisation (LayerNorm), a feed-forward MLP block with another residual connection and LayerNorm. The \texttt{[CLS]} token acts as a global aggregator that attends to all input patches and accumulates information across layers, while skip connections preserve intermediate representations, allowing attention maps to encode both local and global features~\cite{raghu2021vision}. After $L$ layers, the \texttt{[CLS]} token, $Z_\texttt{[CLS]}$, is passed to the classification head to produce the final output. 

Explanation methods for ViTs can be broadly categorised as attention-based and attribution-based approaches.

\textit{Attention-based} explanations leverage the self-attention weights of transformers to highlight regions of the input most influential to the model’s predictions. The attention weights are computed as $A = \text{softmax}(QK^\top/\sqrt{d})$ and the gradients wrt to the attention are computed as $\frac{\partial Z_\texttt{[CLS]}}{\partial A}$. Attention Rollout~\cite{abnar2020quantifying} aggregates attention weights by a series of matrix multiplications across the layers to compute global token importance. Attention Flow~\cite{abnar2020quantifying} propagates attention scores through the network to better capture multi-layer interactions. SAW~\cite{mallick2022saw} scales the attention maps of each layer with the gradient of the attentions with respect to a target class before applying rollout, resulting in class-specific relevance maps.

\textit{Attribution-based} approaches for transformers seek to explain model predictions by propagating relevance or contribution scores from the output back to the input features, typically yielding token or patch level importance maps. Rule-based methods like Layer-wise Relevance Propagation (LRP)~\cite{binder2016layer}, originally proposed for CNNs, modify backpropagation rules to attribute the ``relevance" from the model output back to the inputs. While LRP provides a more structured explanation, it requires careful adaptation of the relevance decomposition rules for different layer types. Initial methods like Grad $\times$ Relevance~\cite{chefer2021transformer}, combined gradients of attention with relevance scores before aggregating by rollout across transformer layers to improve explanations. Subsequent extensions like~\cite{voita2021analyzing, ali2022xai} refined these rules for the non-linear layers and proposed ways to handle numerical instabilities due to relevance propagation through operations such as softmax and LayerNorm. Most recently, AttnLRP~\cite{achtibat2024attnlrp} introduced novel attribution rules for matrix multiplication and previously unhandled operations for transformers to enhance the quality of the explanations.

\section{Methodology} \label{sec:methodology}
Self-attention mechanisms in ViTs naturally lend themselves to use as explanation tools, since it explicitly models dependencies among input tokens. Yet interpreting these models remains challenging because of the many attention heads and layers. Each layer and head contributes differently to the final prediction, and attention is global from the earliest stages, unlike the localised receptive fields in CNNs.
Also, individual attention heads tend to focus on distinct aspects of the input~\cite{lopardo24posthoc}, making head-specific analysis important. These insights motivate our explanation method, which is to retain the ``informative" attentions while filtering out noisy or uninformative signals from specific attention heads. We propose the Rollout Feature Explanation Method (RFEM), which builds on the Attention rollout~\cite{abnar2020quantifying} and CNN-based FEM~\cite{FuadMGBBZ20} methods.
\begin{figure}[]
\centering
\includegraphics[width=\linewidth]{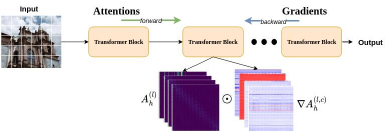} 
\caption{Illustration for RFEM-Class. Gradients with respect to the target class $c$, where $c \in 1, ...,C$, are computed by a backward pass and are used to weight the attention maps at each layer. The remaining steps follow the RFEM method.} \label{fig:rfem_class_framework}
\end{figure}
\vspace{-1mm}
\subsection{Rollout FEM (RFEM)} \label{sec:rfem_desc}
RFEM consists of three key stages:

\textbf{\textit{Layerwise Aggregation}}:
Let the ViT consist of $L$ layers and $H$ attention heads per layer. For each head $h = 1, \dots, H$ and layer $l = 1, \dots, L$, the attention is computed as $A = \text{softmax}(QK^\top/\sqrt{d})$ and we denote the attention matrix as 
\begin{equation}
    A^{(l)}_h \in \mathbb{R}^{(N+1) \times (N+1)}
    \label{eq:attentionmatrix}
\end{equation}

where $N$ is the number of image patches (tokens). It is computed as $N = \frac{\mathcal{H}W}{P^2}$ with $\mathcal{H},W$ the height and width of the image, $P$ is the linear size of the patch and the $+1$ in Eq.~\eqref{eq:attentionmatrix} accounts for the \texttt{[CLS]} token. 

Unlike standard Attention Rollout~\cite{abnar2020quantifying}, we do not average attentions across heads before aggregation. Instead, each head is retained separately to preserve head-specific information. To account for the skip connections, we add the identity matrix $I$:
\begin{equation}
     \hat{A}^{(l)}_h = A_h^{(l)} + I
\end{equation}
The aggregated attention per head is then computed as given by Eq.~\eqref{eq:rfem_layer_agg}. This process preserves head-specific attentions and prevents the differences from being obscured due to averaging.
\begin{equation}\label{eq:rfem_layer_agg}
\hat{A}_{h} = \prod_{l=1}^{L}A_h^{(l)} \quad \text{with } h = 1, \dots, H
\end{equation}

\textbf{\textit{Attention Filtering via Statistical Thresholding}}:
After the aggregation, the attention maps $\hat{A}_h$ often contain dense, low-contrast information. To highlight the more meaningful attention signals, we apply \textit{$K$-sigma thresholding} to each $\hat{A}_h$, following the principle of FEM~\cite{FuadMGBBZ20}, as defined in Eq.~\eqref{eq:k-sigma-thresholding}. This adaptive thresholding uses attention statistics to identify the values that deviate from the mean, treating them as ``informative". Stronger and less frequent attention values are more likely to highlight ``important" regions rather than background information. This intuition is supported by prior work~\cite{chefer2021transformer}, which suggests that sparse, high-contrast attentions often align with salient image features or decision-critical regions.
\begin{equation}
  \bar{A_{h}} = 
     \begin{cases}
      \text{1}, &\quad\text{if $\hat{A}_{h}(i,j) \geq \mu_h + K * \sigma_h$}\\
      \text{0}, &\quad\text{otherwise}
     \end{cases}
     \label{eq:k-sigma-thresholding} 
\end{equation}
For each head $h$, we compute the mean $\mu_h$ and standard deviation $\sigma_h$ of the values in $\hat{A}_h$. An attention value is retained only if it exceeds the threshold $\mu_h + K \cdot \sigma_h$. The hyperparameter $K$ controls how far a value must be from the mean to be considered ``rare" and thus more informative.

Using such a filtering offers the following benefits:

\textbf{ Head-specific adaptation}: The filtering threshold is not fixed and adapts to the attention values per head which is useful as each head attends to different features~\cite{raghu2021vision}.
 
 \textbf{Retains ``Strong" Heads better}: Low-variance heads produce flat attention maps, which, if averaged with others like in Attention Rollout, can dilute strong signals. Our thresholding suppresses values near the mean, preserving only those attentions that are ``different" from the others.

\textbf{Brings sparsity}: Attention-based methods produce dense maps and applying our method helps in highlighting the most salient patches to make it easily understandable for a human viewer.

\textbf{\textit{Head Aggregation}}:
To obtain the final explanation map, we combine the filtered outputs $\bar{A}_h$ from all heads. As discussed in~\cite{mylonas2024attention}, common aggregation operations include a simple average, or selecting the maximum or minimum across the heads. In the earlier work for CNNs~\cite{ayyar2023feature}, a weighted linear combination was proposed with the weights as the average activation values of each map. The choice of the statistic depends on what our objective is: mean weights emphasise overall activity across heads, while max weights highlight the discriminative heads with stronger responses. In our method, we adopt the latter, implementing a weighted linear combination with the maximum value of each $\hat{A}_h$ as the weight as given in Eq.~\ref{eq:rfem_map}
\begin{equation}
A_{\text{rfem}} = \sum_{h=1}^{H} w_h \cdot \bar{A}_h, \quad \text{ where } w_h = \text{max}(\hat{A}_h)
\label{eq:rfem_map}
\end{equation}
The complete RFEM pipeline is illustrated in Figure~\ref{fig:rfem_framework}, using the ViT-B16 architecture with 12 layers and 12 heads per layer as an example. The figure outlines how attention maps are first aggregated, then filtered, and finally combined to produce the RFEM map.

\begin{figure}[]
    \centering
    \begin{subfigure}{0.23\linewidth}
        \centering
        \includegraphics[width=\linewidth]{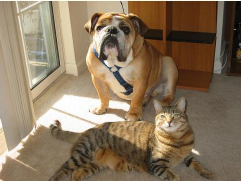}
        \caption{ Image}
    \end{subfigure}
    \begin{subfigure}{0.23\linewidth}
        \centering
        \includegraphics[width=\linewidth]{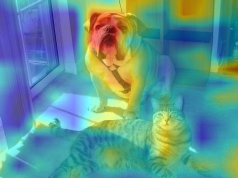}
        \caption{ Rollout}
    \end{subfigure}
    \begin{subfigure}{0.23\linewidth}
        \centering
        \includegraphics[width=\linewidth]{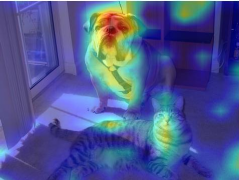}
        \caption{RFEM}
    \end{subfigure}
    \begin{subfigure}{0.23\linewidth}
        \centering
        \includegraphics[width=\linewidth]{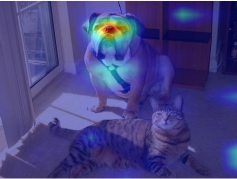}
        \caption{RFEM-C}
    \end{subfigure}
    \caption{For the image (a) our methods (c) and (d) are better than the Attention Rollout (b) map. RFEM filters out the ``unimportant" attentions to focus more on the the main objects and by including the gradient information for RFEM-Class it focuses on the main object in the image.}
    \label{fig:init_results}
\end{figure}

\subsection{RFEM-Class: Incorporating class information} \label{sec:rfem_class}
Similar to Attention rollout, the RFEM method does not consider any information from the output layer and therefore produces class-agnostic explanations. In contrast, other XAI methods for transformers introduce class-specificity by using gradients of the target class $c$ with respect to attention maps~\cite{chefer2021transformer, mallick2022saw}. These methods use the gradients as importance weights to modulate the attentions at each layer, typically averaging across heads and aggregating over layers using the rollout strategy.

We observe that the RFEM pipeline can be seamlessly extended to be class-specific with a minimal modification. Since RFEM filters across heads only after layer aggregation, we propose RFEM-Class, to multiply the attention with its gradients before aggregation and filtering. For each attention head $h$ in layer $l$, we compute the gradient of the attention with respect to the target class $c$, where $c = 1, \dots, C$. This gradient is used to modulate the corresponding attention values before aggregation and an identity matrix is added to account for the skip connections as given by Eq.~\eqref{eq:atttention_and_gradient}. Here, $\odot$ denotes element-wise multiplication.
\begin{equation}
     \hat{A}^{(l,c)}_h = (A_h^{(l)} \odot \nabla A_h^{(l,c)}) + I \label{eq:atttention_and_gradient}
\end{equation}

Similar to RFEM, the attentions are aggregated per head as:
\begin{equation}
    \hat{A}^c_h = \prod_{l=1}^{L} \hat{A}_{h}^{(l,c)} \quad \text{with } h = 1, \dots, H \label{eq:rfem_class_aggregation}
\end{equation}

Following this, the adaptive K-sigma filtering (Eq.~\eqref{eq:k-sigma-thresholding}) is applied to each aggregated map. The filtered attention maps are combined using a linear combination with the maximum attention value per head as the weight, to get the RFEM-Class maps. Figure~\ref{fig:init_results} illustrates the improvements of our methods over the (b) rollout method for image (a). The filtering of RFEM suppresses the noisy background signals seen in rollout, focusing the explanations on the objects in the image. In Figure~\ref{fig:init_results} (d), by incorporating the class-specific information the RFEM-Class map highlights the main object in the image.

\begin{table*}[]
\scriptsize
\centering
\caption{Comparison of explanation methods on the MexCulture~\cite{obeso2018comparative} dataset. Best values in \textbf{bold}, second best in \textit{italics}}
\label{tab:rfem_mexculture}
\resizebox{\linewidth}{!}{%
\begin{tabular}{@{}cllcc|ccccc@{}}
\toprule
& \textbf{} & \textbf{Metric} & \multicolumn{2}{c |}{\textbf{Class Agnostic}} & \multicolumn{5}{c}{\textbf{Class Dependent}} \\
\cmidrule(lr){4-5} \cmidrule(lr){6-10}
& & & RFEM & Rollout & RFEM-Class & SAW & G $\times$ R & AttnLRP  & GradCAM\\
\midrule
\multirow{4}{*}{Plausibility}
& & SIM $\uparrow$   & \textit{0.661 $\pm$ 0.055} & 0.647 $\pm$ 0.053 & \textbf{0.667 $\pm$ 0.080} & 0.655 $\pm$ 0.069 & 0.625 $\pm$ 0.081 & 0.551 $\pm$ 0.084 & 0.567 $\pm$ 0.077 \\
& & PCC $\uparrow$   & 0.450 $\pm$ 0.170 & 0.410.0 $\pm$ 0.173 & \textbf{0.489 $\pm$ 0.167} & \textit{0.458 $\pm$ 0.156} & 0.447 $\pm$ 0.162 & 0.417 $\pm$ 0.155 & 0.388 $\pm$ 0.152 \\
& & AUC $\uparrow$   & \textit{0.773 $\pm$ 0.176} & 0.702 $\pm$ 0.192 & \textbf{0.807 $\pm$ 0.138} & 0.788 $\pm$ 0.124 & 0.752 $\pm$ 0.134 & 0.615 $\pm$ 0.106 & 0.573 $\pm$ 0.198 \\
& & NSS $\uparrow$   & 1.01 $\pm$ 0.74 & 0.791 $\pm$ 0.758 & \textbf{1.13 $\pm$ 0.72} & 0.996 $\pm$ 0.246 & 0.660 $\pm$ 0.901 & 0.730 $\pm$ 0.407 & 0.219 $\pm$ 0.66 \\
\midrule
\multirow{6}{*}{Correctness}
& & IAUC $\uparrow$  & \textit{0.867 $\pm$ 0.089} & 0.848 $\pm$ 0.090 & \textbf{0.878 $\pm$ 0.083} & 0.861 $\pm$ 0.080 & 0.867 $\pm$ 0.0781 & 0.552 $\pm$ 0.445 & 0.820 $\pm$ 0.082 \\
& & DAUC $\downarrow$ & 0.323 $\pm$ 0.169 & 0.345 $\pm$ 0.179 & \textbf{0.282 $\pm$ 0.147} & \textit{0.296 $\pm$ 0.154} & 0.325 $\pm$ 0.155 & 0.315 $\pm$ 0.190 & 0.414 $\pm$ 0.173 \\
& & AD $\downarrow$ & 0.034 $\pm$ 0.10 & 0.045 $\pm$ 0.124 & \textbf{0.013 $\pm$ 0.052} & 0.052 $\pm$ 0.134 & 0.622 $\pm$ 0.175 & 0.082 $\pm$ 0.201 & 0.108 $\pm$ 0.209\\
& & AG $\uparrow$& 0.075 $\pm$ 0.170 & 0.058 $\pm$ 0.147 & \textbf{0.121 $\pm$ 0.220} & 0.101 $\pm$ 0.127 & 0.067 $\pm$ 0.166 & \textit{0.107 $\pm$ 0.221} & 0.0479 $\pm$ 0.134 \\
& & AI $\uparrow$ & 0.489 $\pm$ 0.49 & 0.465 $\pm$ 0.496 & \textbf{0.617 $\pm$ 0.486} & 0.437 $\pm$ 0.50 & 0.486 $\pm$ 0.50 &\textit{0.520 $\pm$ 0.501} & 0.416 $\pm$ 0.492 \\
\midrule
 \multirow{2}{*}{Stability}
& & LIP $\downarrow$ & \textbf{1.76 $\pm$ 1.03} & 2.34 $\pm$ 1.33 & 3.59 $\pm$ 2.6 & 3.79 $\pm$ 2.73 & 13.511 $\pm$  10.60 &  6.110 $\pm$ 5.60 & 6.218 $\pm$ 0.596 \\
& & LSS  $\downarrow$ & \textbf{0.0019 $\pm$ 0.00 } & \textbf{0.0019 $\pm$ 00} & \textbf{0.0019 $\pm$ 0.00}  & \textbf{0.0019 $\pm$ 00} & 0.002 $\pm$ 00  &  0.0021 $\pm$ 0.002 & 0.005 $\pm$ 0.002\\
\bottomrule
\end{tabular}
}
\end{table*}

\begin{table*}[]
\scriptsize
\centering
\caption{Comparison of explanation methods for the samples from SALICON~\cite{jiang2015salicon} dataset. Best values in \textbf{bold}, second best in \textit{italics}.}
\label{tab:rfem_salicon50}
\resizebox{\linewidth}{!}{%
\begin{tabular}{@{}cllcc|ccccc@{}}
\toprule
 &\textbf{} & \textbf{Metric} & \multicolumn{2}{c|}{\textbf{Class Agnostic}} & \multicolumn{5}{c}{\textbf{Class Dependent}} \\
\cmidrule(lr){4-5} \cmidrule(lr){6-10}
 & & & RFEM & Rollout & RFEM-Class & SAW & G $\times$ R & AttnLRP  & GradCAM \\
\midrule
\multirow{4}{*}{Plausibility}
 & & SIM $\uparrow$   & \textit{0.704 $\pm$ 0.05} & 0.647 $\pm$ 0.047 & \textbf{0.716 $\pm$ 0.04} & 0.698 $\pm$ 0.055 & 0.685 $\pm$ 0.056 & \textbf{0.716 $\pm$ 0.066} & 0.673 $\pm$ 0.066 \\
 & & PCC $\uparrow$   & 0.603 $\pm$ 0.09 & 0.550 $\pm$ 0.093 & \textbf{0.676 $\pm$ 0.099} & 0.641 $\pm$ 0.101 & 0.630 $\pm$ 0.106 & \textit{0.673 $\pm$ 0.109}  & 0.630 $\pm$ 0.109 \\
 & & AUC $\uparrow$   &\textit{0.581 $\pm$ 0.12} & 0.57 $\pm$ 0.15 & \textbf{0.610 $\pm$ 0.11} & 0.56 $\pm$ 0.15 & 0.58 $\pm$ 0.15 & 0.494 $\pm$ 0.095 & 0.498 $\pm$ 0.10 \\
 & & NSS $\uparrow$   & 0.205 $\pm$ 0.5 & 0.205 $\pm$ 0.51 &\textbf{0.248 $\pm$ 0.4} & 0.175 $\pm$ 0.51 & \textit{0.225 $\pm$ 0.53} & -0.053 $\pm$ 0.27  & -0.035 $\pm$ 0.37 \\
\midrule
 \multirow{6}{*}{Correctness}
 & & IAUC $\uparrow$  & \textbf{0.518 $\pm$ 0.035} & 0.517 $\pm$ 0.033 & \textbf{0.518 $\pm$ 0.033} &\textbf{ 0.518 $\pm$ 0.035} & \textbf{0.518 $\pm$ 0.035} & 0.515 $\pm$ 0.039 & \textbf{0.518 $\pm$ 0.035} \\
 & & DAUC $\downarrow$ & 0.323 $\pm$ 0.058 & 0.324 $\pm$ 0.059 & \textit{0.321 $\pm$ 0.057} & 0.323 $\pm$ 0.058 & 0.325 $\pm$ 0.057 & \textbf{0.312 $\pm$ 0.063} & 0.324 $\pm$ 0.059 \\
 & & AD $\downarrow$ & \textit{0.037 $\pm$ 0.06} & 0.042 $\pm$ 0.08 & \textbf{0.027 $\pm$ 0.06} & \textit{0.037 $\pm$ 0.065} & \textit{0.037 $\pm$ 0.06} & 0.067 $\pm$ 0.14 & 0.134 $\pm$ 0.13\\
 & & AG $\uparrow$&  0.035 $\pm$ 0.04 & 0.033 $\pm$ 0.042 & \textbf{0.046 $\pm$ 0.043} & 0.042 $\pm$ 0.049 & 0.033 $\pm$ 0.04 & 0\textit{.045 $\pm$ 0.05} & 0.012 $\pm$ 0.027\\
 & & AI $\uparrow$ & 0.617 $\pm$ 0.48 & 0.612 $\pm$ 0.5 & \textbf{0.73 $\pm$ 0.4} & 0.721 $\pm$ 0.44 & 0.59 $\pm$ 0.44 & \textit{0.724 $\pm$0.44} & 0.274 $\pm$ 0.45 \\
\midrule
\multirow{2}{*}{Stability}
 & & LIP $\downarrow$ & \textbf{1.29 $\pm$ 0.788} & \textit{1.53 $\pm$ 1.08} & 2.910 $\pm$ 2.372 & 3.233 $\pm$ 2.384 &10.29 $\pm$ 9.45 & 2.714 $\pm$ 3.23  & 4.169 $\pm$ 3.084\\
 & & LSS  $\downarrow$ & \textbf{0.002 $\pm$ 0.001} & 0.0026 $\pm$ 0.0011 & 0.0025 $\pm$ 0.0011 & 0.0026 $\pm$ 0.001 & 0.0032 $\pm$ 0.0025 & \textit{0.002 $\pm$ 0.002} & 0.003 $\pm$ 0.0012 \\
\bottomrule
\end{tabular}
}
\end{table*}

\section{Experiments and Results}\label{sec:exp_results}

We compare our method against a set of five representative XAI methods for transformers, including attention-based methods, gradient-based methods, and the latest transformer-specific variant of LRP, AttnLRP~\cite{achtibat2024attnlrp}. All experiments were done on a single NVIDIA A40 46GB GPU. For ImageNet, we use pretrained weights for analysis, while training details of the ViT models for the other two datasets are provided in the Supplementary Materials.

Most XAI evaluations rely on perturbation-based metrics, which measure the importance of input tokens or patches by observing changes in model outputs when these regions are modified~\cite{petsiuk2018rise}. While such metrics assess faithfulness, they do not guarantee human interpretability. To address this limitation, following the method first proposed in \cite{BourrouxBBG22}, we complement standard faithfulness evaluations with experiments on two datasets that provide human-centric ground truth explanations in the form of gaze fixation density maps (GFDMs) and saliency maps.

\subsection{Dataset Details}\label{sec:dataset_details}
Below we describe datasets with human perception recordings as Gaze Fixation Density Maps. 
%We evaluate explanation maps against human gaze fixation density maps (GFDMs) as ground truth on two image datasets, benchmarking XAI methods with respect to human visual perception. In addition, we conduct standard perturbation-based evaluation using random 5000 images from the ImageNet-1K validation dataset.

\textbf{MexCulture}~\cite{obeso2018comparative}: This dataset comprises 20000 images of architectural structures for the classification of cultural heritage buildings in Mexico. It includes four classes: Colonial, Prehispanic, Modern, and Other. The dataset is split into 12000 training, 4000 validation, and 4000 testing images. Additionally, human gaze fixation data and the corresponding GFDMs are publically available for 284 images at\footnote{\url{https://www.nakala.fr/data/11280/5712e468}}. These fixations were collected through a psycho-visual experiment, where participants viewed historical buildings to identify their architectural styles. The GFDMs were generated using the methodology described in~\cite{obeso2018comparative}.
    
\textbf{SALICON}~\cite{jiang2015salicon}: This dataset comprises saliency maps for 20000 images, collected by mouse-pointing using the paradigm that gaze and pointing express the same perception. The dataset is split into 10000 training images, 5000 for validation and 5000 for testing. SALICON is a challenging dataset as it comprises multiple objects in the scene. For our experiments, we randomly select 1000 images from the validation set to evaluate our explanation methods.
% This subset allows us to compute various metrics while keeping GPU processing time manageable.

In addition, we conduct standard perturbation-based evaluation using 5000 random images from the ImageNet1K validation dataset.

\subsection{Metrics}\label{sec:metrics}
We evaluate XAI methods along three dimensions: Correctness (Faithfulness), Stability, and Plausibility. Detailed definitions of all metrics are provided in the Supplementary Materials~\ref{sec:metrics_appendix}.

\textit{Correctness} metrics measure whether the pixels identified as important by the XAI method truly drive model predictions. This is tested through perturbation-based evaluations, where pixels are inserted or removed according to their explanation scores and the effect on predictions is measured. Metrics include insertion AUC (IAUC), deletion AUC (DAUC), average gain (AG), average increase (AI), and average drop (AD). To improve robustness, \cite{blucher2024decoupling} also proposed the $\Delta A^F$ metric, which contrasts two orders (most vs. least relevant first) for the deletion perturbation. \textit{Stability} metrics evaluate whether small input changes lead to large variations in explanations. The Lipschitz criterion (LIP) quantifies sensitivity to perturbations, and local surrogate stability (LSS) checks the consistency of the explanation under neighbourhood approximations. A detailed methodology for evaluation with this metric is presented in~\cite{xu2023stability}. \textit{Plausibility} metrics measure alignment with human perception. Since this is crucial in domains such as medical imaging, we compare saliency maps with human gaze data and report Similarity (SIM), Pearson’s correlation (PCC), Normalized scanpath saliency (NSS), and AUC-Judd metrics.

\subsection{Results}
We compare RFEM and RFEM-Class with five transformer-based XAI methods: the state-of-the-art AttnLRP~\cite{achtibat2024attnlrp}, Attention Rollout~\cite{abnar2020quantifying}, SAW~\cite{mallick2022saw}, Grad $\times$ Relevance~\cite{chefer2021transformer}, and GradCAM~\cite{selvaraju2017grad}, adapted for transformer models. All methods are quantitatively evaluated using the metrics described in the previous section. We further conduct qualitative analyses by overlaying the XAI maps on dataset images and report both the theoretical complexity and empirical runtimes of our methods. For consistency, all images and XAI maps are resized to $224 \times 224$ for visualization. 

\subsubsection{Quantitative Evaluation}
Tables~\ref{tab:rfem_mexculture} and ~\ref{tab:rfem_salicon50} present the values of the metrics for the different XAI methods. Across both the MexCulture and SALICON datasets, we observe that our RFEM and RFEM-Class methods show consistently strong performances for plausibility, correctness, and stability metrics. The class-dependent RFEM-Class method, systematically scores the highest or second-highest in \textit{plausibility} metrics such as SIM, PCC, AUC, and NSS, reflecting its ability to produce explanations that align well with ground-truth human attention patterns. RFEM (class-agnostic) also has high plausibility values, making it also a reliable choice. This highlights that RFEM-Class uses class-aware guidance to refine the localization of salient regions, whereas RFEM produces robust, general-purpose explanations.

Across the \textit{correctness} metrics for the MexCulture dataset, RFEM-Class consistently outperformed all other XAI methods, indicating that its explanations are more faithful to the model’s internal decision process. On the SALICON dataset, RFEM-Class and RFEM achieve either the best or the second-best to AttnLRP, while SAW and Grad $\times$ Relevance occasionally reach this performance on isolated metrics (e.g., IAUC). Notably, the Rollout method, trails behind our RFEM methods in nearly all metrics, suggesting it captures broad attention trends but lacks fine-grained attribution quality. Thus, the proposed refining of the attentions through statistical filtering indeed produces better explanations. 

The AUC and NSS metrics for all methods are lower on the SALICON dataset, likely due to the presence of multiple objects in an image, compared to the single object images in the MexCulture dataset. However, the higher SIM and PCC scores indicate that our methods are able to capture the overall spatial distribution of the saliency. The SOTA AttnLRP is better on some metrics on the SALICON dataset like PCC and DAUC. However, this performance is not consistent across the two datasets. If it was best or second best for SALICON, it is ranked below the other methods for plausibility on MexCulture. The same could be said for the stability metric LIP where it was stable for SALICON but not MexCulture. In contrast, RFEM and RFEM-Class are relatively close in most metrics to AttnLRP when it has the best performance and have consistent scores across the two datasets. 

\begin{figure*}[] % span both columns, remove * if single column
    \centering
    \scalebox{0.8}{ % shrink uniformly to fit
    \begin{tabular}{c *{9}{c}}
        & {Original} & {GFDM} & {GradCAM} & {AttnLRP} & {$G \times R$} & {Rollout} & {SAW} & {RFEM} & {RFEM-Class} \\ 

        \rotatebox{90}{{Colonial}} &
        \includegraphics[width=0.09\textwidth]{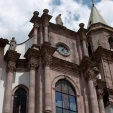} &
        \includegraphics[width=0.09\textwidth]{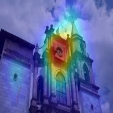} &
        \includegraphics[width=0.09\textwidth]{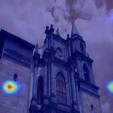} &
        \includegraphics[width=0.09\textwidth]{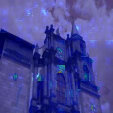} &
        \includegraphics[width=0.09\textwidth]{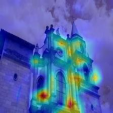} &
        \includegraphics[width=0.09\textwidth]{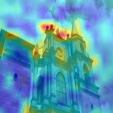} &
        \includegraphics[width=0.09\textwidth]{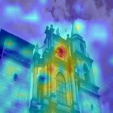} &
        \includegraphics[width=0.09\textwidth]{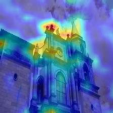} &
        \includegraphics[width=0.09\textwidth]{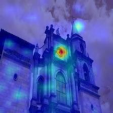} \\

        \rotatebox{90}{{Prehispanic}} &
        \includegraphics[width=0.09\textwidth]{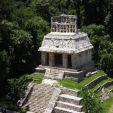} &
        \includegraphics[width=0.09\textwidth]{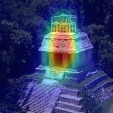} &
        \includegraphics[width=0.09\textwidth]{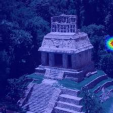} &
        \includegraphics[width=0.09\textwidth]{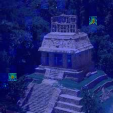} &
        \includegraphics[width=0.09\textwidth]{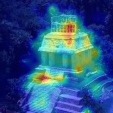} &
        \includegraphics[width=0.09\textwidth]{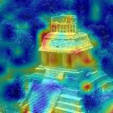} &
        \includegraphics[width=0.09\textwidth]{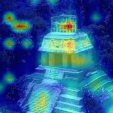} &
        \includegraphics[width=0.09\textwidth]{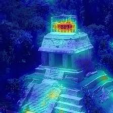} &
        \includegraphics[width=0.09\textwidth]{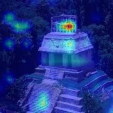} \\

       \rotatebox{90}{{Modern}} &
        \includegraphics[width=0.09\textwidth]{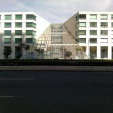} &
        \includegraphics[width=0.09\textwidth]{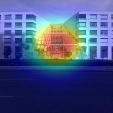} &
        \includegraphics[width=0.09\textwidth]{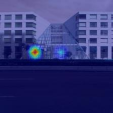} &
        \includegraphics[width=0.09\textwidth]{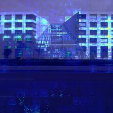} &
        \includegraphics[width=0.09\textwidth]{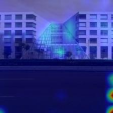} &
        \includegraphics[width=0.09\textwidth]{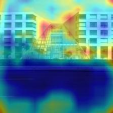} &
        \includegraphics[width=0.09\textwidth]{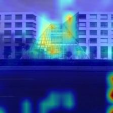} &
        \includegraphics[width=0.09\textwidth]{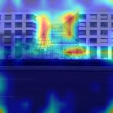} &
        \includegraphics[width=0.09\textwidth]{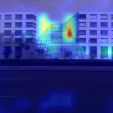} \\
    \end{tabular}
    }
    \caption{Qualitative comparison of the XAI methods across the three classes for MexCulture images}
    \label{fig:mexculture_visualization}
\end{figure*}

\begin{figure*}
\centering
\includegraphics[width=0.9\textwidth]{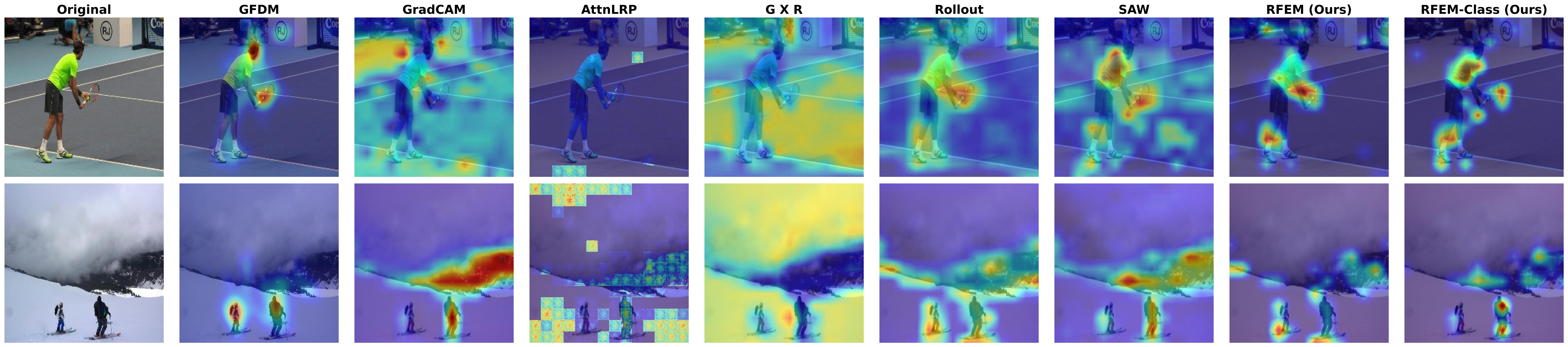} 
\caption{Qualitative comparison of the XAI methods and the GFDMs on SALICON images} \label{fig:salicon50_visualization}
\end{figure*}

\begin{table}[]
\scriptsize
\centering
\caption{Perturbation test results for $\Delta A^F$.\textbf{Bold}: best values, \textit{Italics}: second best}\label{tab:rfem_perturbation_results}
\resizebox{\columnwidth}{!}{%
\begin{tabular}{@{}lccc@{}}
\toprule
\textbf{Methods} & \textbf{ImageNet1K} $\uparrow$ & \textbf{MexCulture} $\uparrow$ & \textbf{SALICON} $\uparrow$ \\
\midrule
Random & 0.02 $\pm$ 0.4 & 0.005 $\pm$ 0.1 & 0.002 $\pm$ 0.03 \\
CB-CAM (\textit{baseline}) & 1.26 $\pm$ 1.20 &0.133 $\pm$ 0.47 &0.013 $\pm$ 0.11 \\
\midrule
GradCAM                         & 1.31 $\pm$ 2.48           & 0.07 $\pm$ 0.36                   &-0.074 $\pm$ 0.07\\
Rollout                        & 7.90 $\pm$ 1.41           & 0.49 $\pm$ 0.8                    & 0.066 $\pm$ 0.12\\
Grad $\times$ Relevance          & 7.09 $\pm$ 1.3            & 0.68 $\pm$ 1.07                   &0.028 $\pm$ 0.1\\
SAW                                   & 7.82 $\pm$ 1.55           & 0.58 $\pm$ 0.89                   & 0.072 $\pm$ 0.12\\
AttnLRP                         & \textbf{8.69 $\pm$ 1.65 }          & \textit{0.90 $\pm$ 1.90}          & \textit{0.26 $\pm$ 0.05}\\
\midrule
RFEM                                                        & 8.05 $\pm$ 1.37           & 0.64 $\pm$ 1.04                   & 0.075 $\pm$ 0.10\\
RFEM-Class                                                 & \textit{8.331 $\pm$ 1.19}           &\textbf{0.92 $\pm$ 1.4}           & \textbf{0.3 $\pm$ 0.14}\\
\bottomrule
\end{tabular}%
}
\end{table}

We further evaluated the methods on 5,000 randomly selected images from the ImageNet1K validation set using the $\Delta A^F$ metric, as shown in Table~\ref{tab:rfem_perturbation_results}. To establish baselines for the perturbation metrics, we additionally compared against a Random saliency map and the Centre-Biased CAM (CB-CAM), as proposed in ~\cite{xu2023stability}, constructed by upsampling a $7 \times 7$ map containing a single 1 at the centre and 0 elsewhere to the image resolution. Since standard classification datasets typically contain a single object near the image centre, CB-CAM provides a meaningful baseline. As reported in Table~\ref{tab:rfem_perturbation_results}, RFEM-Class consistently demonstrates strong performance across datasets, achieving the highest $\Delta A^F$ on both MexCulture and SALICON, and ranking second only to AttnLRP on ImageNet1K. Similarly, RFEM attains competitive $\Delta A^F$ scores on ImageNet1K, outperforming several methods such as SAW, Rollout, and Grad $\times$ Relevance, despite being class-agnostic. Taken together, these perturbation-based evaluations indicate that RFEM and RFEM-Class not only generate human-plausible maps, but also successfully highlight model-relevant regions across diverse datasets.

\subsubsection{Qualitative Evaluation} \label{sec:qualitative_evalaution}
Figures~\ref{fig:mexculture_visualization} and ~\ref{fig:salicon50_visualization} illustrate samples from the MexCulture and SALICON datasets, respectively. For both datasets, we present the original image, the ground-truth GFDM, and the corresponding explanations generated by the XAI methods. Across both datasets, we observe that RFEM and RFEM-Class consistently highlight regions that closely align with human visual attention, as indicated by the GFDMs. In particular, both methods accurately identified the peak fixation locations while suppressing scattered attentions in the image. Their maps are less noisy when compared to Rollout and SAW methods, producing more focused explanations. While AttnLRP also highlights meaningful regions, its maps do not match human attention as well and are more difficult to interpret, exhibiting a checkerboard pattern due to scoring each token individually. Additional visualisations on ImageNet samples and the effect of $K$ for our statistical filtering are presented in the Supplementary Materials~\ref{sec:suppl_res}.

\subsubsection{Computational Complexity and Processing times}\label{sec:computational_analysis}
Consider a ViT with $L$ layers, $H$ attention heads per layer, and $N$ input tokens including $\texttt{[CLS]}$. Let $A^{(l)}_h \in \mathbb{R}^{N \times N}$ denote the attention matrix of head $h$ in layer $l$. Following~\cite{achtibat2024attnlrp}, we measure complexity relative to a single forward pass, treated as $\mathcal{O}(1)$. Since a backward pass is comparable, we approximate it as $\mathcal{O}(1)$ as well. As summarised in Table~\ref{tab:xai_complexity_analysis}, methods that aggregate attention maps (Rollout, SAW, RFEM and RFEM-Class) operate on $L \times H$ attention matrices of size $N \times N$, yielding $\mathcal{O}(LHN^2)$ complexity. In contrast, as AttnLRP and Grad $\times$ Relevance require only forward and backward passes and have a theoretical complexity $\mathcal{O}(1)$.

\begin{table}[]
\scriptsize
\centering
\caption{Theoretical computational complexity of XAI methods for ViT. A forward pass is $\mathcal{O}(1)$. Mult: multiplications, Grad: gradient of attentions}
\resizebox{0.95\columnwidth}{!}{%
\begin{tabular}{@{}lcccc@{}}
\toprule
\textbf{Methods} & $\rightarrow$ & $\leftarrow$  & \textbf{Main Operation} & \textbf{Complexity} \\
\midrule
AttnLRP    & 2 & 1 & Full passes & $\mathcal{O}(1)$ \\ 
Grad $\times$ Relevance & 2& 1& Full passes & $\mathcal{O}(1)$ \\
GradCAM   & 1 & 1 & $A_L$ $\times$ gradient & $\mathcal{O}(H \cdot N^2)$ \\ 
Rollout    & 1 & 0 & Head avg + layer mult & $\mathcal{O}(L \cdot H \cdot N^2)$ \\ 
SAW        & 1 & 1 & Grad $\times$ Attn  & $\mathcal{O}(L \cdot H \cdot N^2)$ \\ 
RFEM       & 1 & 0 & Per-head Mult & $\mathcal{O}(L \cdot H \cdot N^2)$ \\ 
RFEM-Class & 1 & 1 & Grad $\times$ Attn & $\mathcal{O}(L \cdot H \cdot N^2)$ \\ 
\bottomrule
\end{tabular}}
\label{tab:xai_complexity_analysis}
\end{table}

However, theoretical complexity does not fully capture runtime efficiency. In practice, a single backward pass in a ViT costs roughly twice as much as a forward pass~\cite{gpuprofile}. As shown in Table~\ref{tab:runtime}, Grad $\times$ Relevance and AttnLRP that backpropagate the relevance from the output to the input are empirically the slowest. Rollout and RFEM, which each require only a single forward pass, are the fastest. GradCAM, SAW and our RFEM-Class compute attention gradients and are slower. RFEM-Class emerged as the third fastest method overall. These results confirm that the filtering operations in RFEM and RFEM-Class introduce minimal overhead, in practice.

\begin{table}[]
\small
\centering
\caption{Per-image runtime in seconds comparison for different XAI methods on the ViT-base16 model}
\begin{tabular}{@{}lc@{}}
\toprule
\textbf{Method} & \textbf{Per image Runtime (s)} \\
\midrule
AttnLRP       & $0.09 \pm 0.014$ \\
Grad $\times$ Relevance & $0.182 \pm 0.008$ \\
GradCAM       & $0.032 \pm 0.004$ \\
Rollout       & $0.011 \pm 0.001$ \\
SAW           & $0.043 \pm 0.006$ \\
RFEM          & $0.012 \pm 0.0014$ \\
RFEM-Class    & $0.030 \pm 0.002$ \\
\bottomrule
\end{tabular}
\label{tab:runtime}
\end{table}

\section{Conclusion and Perspectives} \label{sec:conclusion}
In this work, we proposed a ViT explanation method RFEM, based on the statistical filtering of attention heads. By statistically thresholding the attention maps, RFEM removed uninformative values and highlighted rare yet informative ones. We further introduced RFEM-Class, which combined attention gradients to provide class-specific explanations. Both methods achieved high plausibility scores with respect to human perception ground truth on MexCulture and SALICON datasets and performed competitively against representative XAI methods on standard faithfulness metrics across these datasets and ImageNet. Overall, our results show that attention could serve as reliable explanations, offering a faster alternative to attribution methods while remaining human interpretable. A potential future work would be to analyse our methods for other modalities.
%\subsubsection*{References}

\clearpage
\appendix
\thispagestyle{empty}

% Supplementary material: To improve readability, you must use a single-column format for the supplementary material.
\onecolumn
\aistatstitle{There is  More to Attention: Statistical Filtering Enhances Explanations in Vision Transformers: \\
Supplementary Materials}

% Note: You can choose whether the include you appendices as part of the main submission file (here) OR submit them separately as part of the supplementary material. It is the authors' responsibility that any supplementary material does not conflict in content with the main paper (e.g., the separately uploaded additional material is not an updated version of the one appended to the manuscript).

\section{METRICS}\label{sec:metrics_appendix}
Among the different ways of evaluating an XAI method, we draw inspiration from ~\cite{xu2023stability} to evaluate the \textit{Correctness}, \textit{Plausibility} and \textit{Stability} of the explanation maps. Correctness (faithfulness) quantifies the degree to which an explanation reflects the model’s internal decision-making. Plausibility (coherency) assesses whether the explanation is meaningful and interpretable from a human perspective. Stability (sensitivity) measures the consistency of explanations, evaluating whether similar inputs yield similar explanation maps.

\subsection{Correctness}
\label{subsec:correctness}

\begin{itemize}
    \item \textbf{Deletion Area Under Curve (DAUC)}~\cite{petsiuk2018rise}: It iteratively masks out the important regions of an input image, identified by the explanation map $S$, from the most to the least important pixels. It then measures the variations in the model output with these masked inputs and the area under this curve gives the DAUC curve. In our experiments, we use a step size of 50 to uniformly mask pixels, progressing from no masking to complete masking. For a faithful explanation method, we expect a rapid drop in the model output as the most important regions are masked, resulting in a lower DAUC value.
    
    \item \textbf{Insertion Area Under Curve (IAUC)}~\cite{petsiuk2018rise}:This metric complements DAUC by starting with an empty image and progressively adding pixels based on the importance scores from the explanation map $S$. For a faithful explanation method, the model output should increase rapidly as the most important regions are added, resulting in a higher IAUC value.
    
    \item \textbf{Average Drop (AD)}~\cite{chattopadhay2018grad}: It measures the average decrease in model confidence when the input image is masked by the  explanation map $S$. A lower AD indicates a more faithful saliency map as given by Equation~\eqref{eq:AD}, where $p_c^i$ is the original model output for a class $c$, $o_c^i$ is the confidence for the masked input.
    \begin{equation}\label{eq:AD}
        \text{AD} = \frac{1}{N} \sum_{i=1}^{N} \frac{max (0, p_c^i - o_c^i)}{p_c^i}
    \end{equation}
    
    \item \textbf{Average Increase (AI)}~\cite{chattopadhay2018grad}: It measures the fraction of test samples for which the model's confidence increases when it is only given the input masked by the explanation map as given by Equation~\eqref{eq:AI}. A higher AI suggest that the explanation method highlights informative regions. 
    \begin{equation}
        \label{eq:AI}
        \text{AI} = \frac{1}{N}\sum_{i=1}^{N} \mathbbm{1} \cdot (o_c^i > p_c^i)
    \end{equation}
    \item \textbf{Average Gain (AG)}~\cite{zhang2024opti}: It is the symmetric counterpart for AD and measures the gain in class probability after masking the input with the explanation map as given by Equation~\eqref{eq:AG}. A higher AG indicates that the XAI method identifies important regions and also enhances the \textit{certainty} of the model. 
    \begin{equation}
        \label{eq:AG}
        \text{AG} = \frac{1}{N} \sum_{i=1}^{N} \frac{max (0,  o_c^i - p_c^i)}{1 - p_c^i}
    \end{equation}
\end{itemize}

There exist multiple ways to measure perturbation-based faithfulness. Recent works introduced refined metrics to better account for out-of-distribution effects or baseline sensitivity~\cite{samek2016evaluating, blucher2024decoupling}. DAUC and IAUC ~\cite{petsiuk2018rise} remain widely adopted benchmarks, measuring how model output changes as pixels are progressively added to or removed from the image based on the saliency rankings. Recent evaluations extend this by considering both most relevant first (MoRF) and least relevant first (LeRF) perturbations orders. Combined with deletion (also called flipping (F)) and insertion (I), this yields four AUC metrics: $A^{F}_{\text{MoRF}}$, $A^{F}_{\text{LeRF}}$, $A^{I}_{\text{MoRF}}$ and $A^{I}_{\text{LeRF}}$~\cite{achtibat2024attnlrp, blucher2024decoupling}. To further enhance robustness, the $\Delta A$ score is often used, which is computed as the difference between area under the LeRF and MoRF curves, offering a more stable measure of explanation quality.

\subsection{Plausibility}
In this work, we compare explanation maps generated by different XAI methods against the GFDMs provided with the datasets, which we treat as ground truth. The plausibility of an explanation map is defined as its correspondence with human visual perception of the content. The closer an XAI map is to the GFDM for a given image, the more effective the sample-based explainer is. For this comparison, we employ established saliency and explanation map evaluation metrics, as suggested in~\cite{wang2017deep}. Since XAI maps are ultimately intended for human interpretation, we argue that plausibility metrics should take precedence over alternative evaluation criteria, as they directly assess human interpretability while still providing insight into the model’s decision-making process.

\begin{itemize}
    \item \textbf{Similarity (Sim)}: It measures the similarity between the GFDM maps $G$ and the explanation map $S$ by treating them as discrete probability distributions, as defined in Equation~\eqref{eq:sim}. A SIM value of 0 indicates no overlap between the distributions, whereas a SIM value of 1 indicates identical distributions.
    \begin{equation}
        \label{eq:sim}
        \mathit{SIM}(S, G) = \sum_{i} \mathit{min} (S^{'}_{i}, G^{'}_{i})
        \quad \text{where} \quad
\sum_{i} S'_{i} = 1 \quad \text{and} \quad \sum_{i} G'_{i} = 1
    \end{equation}
    
    \item \textbf{Pearson Correlation Coefficient (PCC)}: It measures the strength and direction of the linear relationship between the GFDM $G$ and the explanation $S$, treated as random variables, as shown in Equation~\eqref{eq:pcc}.  Here, $\mathrm{cov}(S,G)$ is the covariance of $S$ and $G$, while $\sigma_{S}$ and $\sigma_{G}$ are the standard deviations. PCC of 1 indicates perfect correlation, 0 indicates no correlation and a value close to -1 indicates strong negative correlation. 
    
        \begin{equation}
        \label{eq:pcc}
        \mathit{PCC}(S, G) = \frac{\mathrm{cov}(S, G)}{\sigma_{S} \times \sigma_{G}}
    \end{equation}

    \item \textbf{AUC-Judd}: This metric~\cite{wang2017deep} computes the area under the Receiver Operating Characteristic (ROC) curve and is widely adopted for evaluating saliency maps. Given a GFDM map 
$G$, fixation locations are defined by thresholding $G$ to obtain a binary fixation map $F$, where pixels with fixation values are marked as positives. The explanation map 
$S$ is then treated as a prediction map, and each pixel is classified as fixation or non-fixation with respect to 
$F$. By varying a decision threshold over the continuous values of 
$S$ from 0 to 1, one can compute the corresponding true positive rate (TPR) and false positive rate (FPR), thereby generating the ROC curve. The AUC-Judd score is defined as the area under this curve, with higher values indicating stronger alignment between the explanation map and human fixation patterns. We select the top-5\% of the pixels from the GFDM $G$ for $F$.
    
        \item \textbf{Normalized Scanpath Saliency (NSS)}: Similar to AUC, NSS also first creates the map of fixations $F$ by thresholding $G$ where $N$ is the total fixation points in $F$.
        It measures how strong the saliency is at the "important" fixation locations, as given in Equation~\eqref{eq:nss}, where $\mu(.)$ and $\sigma(.)$ are mean and standard deviation respectively
        
        \begin{equation}
        \label{eq:nss}
            \text{NSS} = \frac{1}{N}\sum_{i=1}^{N} \bar{S} \times F(i), \text{   where    } \bar{S} = \frac{S - \mu (S)}{\sigma (S)}
        \end{equation}
\end{itemize}

\subsection{Stability}
Understanding the stability of explanation methods is critical when evaluating their reliability. XAI should produce similar explanations for similar inputs, especially in the local neighbourhood of a given input sample.
\begin{itemize}
    \item \textbf{Lipschitz criterion (LIP)}: The LIP metric~\cite{alvarez2018towards} draws inspiration from Lipschitz continuity criterion to measure the local stability of the explanation method. It measures the stability of the method, by evaluating the maximum ratio of change in explanations to the change in inputs within a small radius $\epsilon$ as shown in Equation~\eqref{eq:LIP}. Here $X_0$ is original input, $\widetilde{X}$ is the perturbed version of the input and $S_X$ is the explanation map for an input $X$. 

    \begin{equation} \label{eq:LIP}
        LIP(X_0) = \text{max}_{\Vert X_0 - \widetilde{X}\Vert_2 < \epsilon} \frac{\Vert S_{X_0} - S_{\widetilde{X}} \Vert_2}{\Vert X_0 - \widetilde{X}\Vert_2} 
    \end{equation}
 However, this metric does not take into account the underlying model behaviour as an unstable model output for $X_0$ would result in a corresponding unstable $S_{X_0}$ which should be a correct explanation but LIP would penalise this. Thus, ~\cite{xu2023stability} propose the Local Surrogate Stability (LSS)  metric. 

    \item \textbf{Local Surrogate Stability (LSS)}: Unlike LIP, LSS incorporates model behaviour while measuring explanation stability under assumption that explanations should vary only when the model's behaviour changes. To evaluate the local behaviour, a surrogate model is constructed from the explanation method as proposed by~\cite{yeh2019fidelity}. This surrogate serves as a first-order linear approximation of the model’s output, using the saliency map at $X_0$ and is defined as:
    \[
    E_{X} = S_{X_0}^\top \dot (X-X_0) + g(X_0) 
    \]
where $S_{X_0}$ is the explanation for a sample $X_0$, $g(X_0)$ is the output (logit) of the model for $X_0$ and $E_{X}$ is surrogate model's approximation of $g(X)$ near $X_0$. This approximation assumes that the model $g$ behaves approximately linearly in a small neighbourhood of $\epsilon$ around $X_0$.
    
    Given $X_0$ and its perturbed version $\widetilde{X}$ such that $\Vert X_0 - \widetilde{X} \Vert_2 < \epsilon$, let $m = \frac{X_0 + \widetilde{X}}{2}$ be their midpoint. The deviation in the explanation at this midpoint is defined as:
\[
D(X_0, \widetilde{X}) = E_{X_0}(m) - E_{\widetilde{X}}(m)
\]
Then the LSS is measured as:
\[
LSS(X_0) =  \text{max}_{\Vert X_0 - \widetilde{X}\Vert_2 < \epsilon} \frac{D(X_0, \widetilde{X})}{\Vert X_0 - \widetilde{X}\Vert_2} 
\]

LSS penalises explanations that change when the model's behaviour does not and rewards explanation methods that only vary when the model's output also varies. 
\end{itemize}

\section{Implementation Details}\label{sec:implementatioN_dets}
All experiments are done using a ViT-B16 model with  12 transformer layers ($L=12$), 12 attention heads ($H=12$), linear patch size of 16 ($P=16$), and an input image resolution of $224 \times 224$. It is initialised with pretrained ImageNet1K weights. For the MexCulture dataset, we fine-tune the model for 20 epochs, with a batch size of 16, achieving a final test accuracy of $89.71\%$. For the SALICON dataset, the model is trained for 25 epochs with a batch size of 32 and reaches a final test accuracy of $82.84\%$. Both the networks were trained using the SGD optimizer with a learning rate of $1e-4$ and an exponential decay of $1e-4$. For experiments involving the ImageNet1K dataset, we use the pretrained model directly. 

We use implementation from the official repository\footnote{\url{https://github.com/hila-chefer/Transformer-Explainability}} for the Attention rollout, Grad $\times$ Relevance and GradCAM methods and is under the MIT License. AttnLRP results were obtained using the lxt library from the authors\footnote{\url{https://github.com/rachtibat/LRP-eXplains-Transformers}} and has the The Clear BSD License. We implemented the code for SAW and our methods. 

\section{ADDITIONAL EXPERIMENTS and RESULTS}\label{sec:suppl_res}

\subsection{Effect of the $K$ Parameter in $K$-sigma rule}
\begin{figure}
\centering
\includegraphics[width=\textwidth]{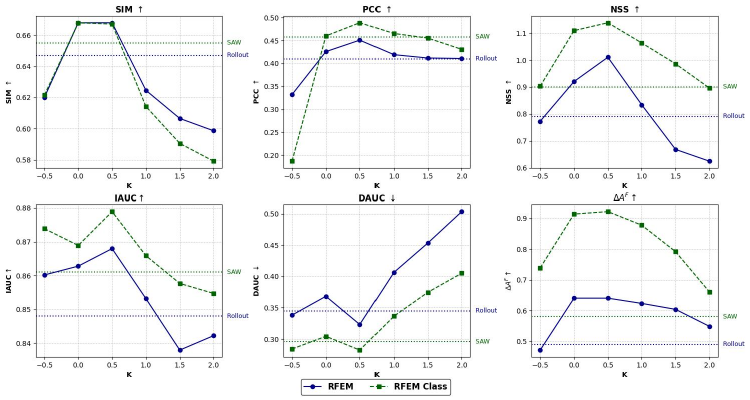} 
\caption{Effect of changing $K = [-0.5, 0, 0.5, 1, 1.5, 2]$ for our methods RFEM and RFEM-Class for the XAI metrics : SIM, PCC, NSS, IAUC, DAUC and $\Delta A^F$ on the MexCulture dataset. The dotted lines represent the corresponding metric values of the baseline methods, Rollout and SAW, from which RFEM and RFEM-Class are derived.} \label{fig:meculture_ablation_visualization}
\end{figure}

we evaluate the class-aware RFEM-Class method against its ablated variant, the class-agnostic RFEM. We analyze both methods across a range of threshold values $K = [-0.5, 0, 0.5, 1, 1.5, 2]$, and include Rollout and SAW metrics as baselines, since RFEM and RFEM-Class are built from these respective approaches. The inclusion of $K = -0.5$ is to examine whether retaining attention values below the mean provides useful information. We used the MexCulture dataset and evaluate multiple metrics, including SIM, PCC, NSS, AUC, IAUC, DAUC, and $\Delta A^F$. As illustrated by Figure~\ref{fig:meculture_ablation_visualization}, RFEM-Class achieves higher perturbation scores ($\Delta A^F$) and lower deletion scores (DAUC) compared to RFEM, highlighting the advantage of incorporating class-awareness for identifying more discriminative regions.

 Across most metrics, thresholding with a moderate $K$ in the range of $0.5$ to $1.0$ is better than Rollout and SAW and extreme values of $K$. For both RFEM and RFEM-Class, SIM, PCC, NSS, and $\Delta A^F$ generally peak in this range, indicating better alignment to ground truth, and improved localization of the salient regions. At very low ($K=-0.5$) or high ($K=1.5$, $2.0$) threshold values, performance degrades across all metrics, suggesting that overly lenient filtering introduces noise while aggressive thresholding discards essential information. When compared to Rollout and SAW, both RFEM and RFEM-Class with statistical filtering are better for both the $\Delta A^F$ and Deletion (DAUC) metrics, thus improving faithfulness. This demonstrates that our methods can enhance the quality and reliability of visual explanations beyond standard attention methods.

\begin{figure}[]
    \centering
    {\scriptsize
    \begin{tabular}{c c c c c}

        \rotatebox{90}{\textbf{Original}}
        & \includegraphics[width=0.15\textwidth]{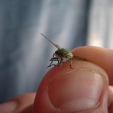} 
        & \includegraphics[width=0.15\textwidth]{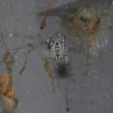} 
        & \includegraphics[width=0.15\textwidth]{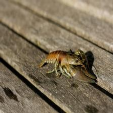}
        & \includegraphics[width=0.15\textwidth]{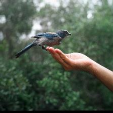}\\

        \rotatebox{90}{{\textbf{GradCAM}}}
        & \includegraphics[width=0.15\textwidth]{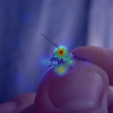} 
        & \includegraphics[width=0.15\textwidth]{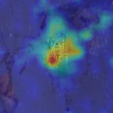} 
        & \includegraphics[width=0.15\textwidth]{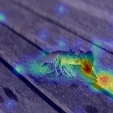} 
        & \includegraphics[width=0.15\textwidth]{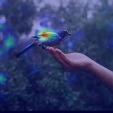}        \\
        
        \rotatebox{90}{{\textbf{AttnLRP}}}
        & \includegraphics[width=0.15\textwidth]{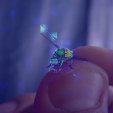} 
        & \includegraphics[width=0.15\textwidth]{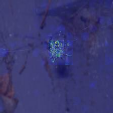} 
        & \includegraphics[width=0.15\textwidth]{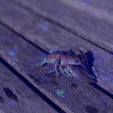}
        & \includegraphics[width=0.15\textwidth]{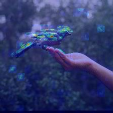}\\

        \rotatebox{90}{\textbf{{Grad $\times$ Relevance}}}
        & \includegraphics[width=0.15\textwidth]{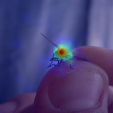} 
        & \includegraphics[width=0.15\textwidth]{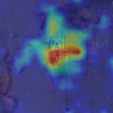} 
        & \includegraphics[width=0.15\textwidth]{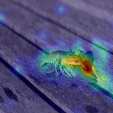}
        & \includegraphics[width=0.15\textwidth]{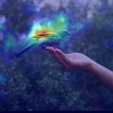}\\

        \rotatebox{90}{{\textbf{Rollout}}}
        & \includegraphics[width=0.15\textwidth]{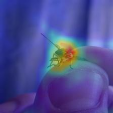} 
        & \includegraphics[width=0.15\textwidth]{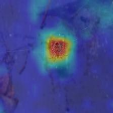} 
        & \includegraphics[width=0.15\textwidth]{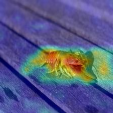}
        & \includegraphics[width=0.15\textwidth]{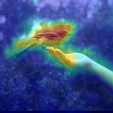}\\

        \rotatebox{90}{{\textbf{SAW}}}
        & \includegraphics[width=0.15\textwidth]{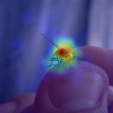} 
        & \includegraphics[width=0.15\textwidth]{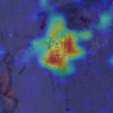} 
        & \includegraphics[width=0.15\textwidth]{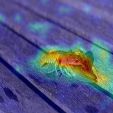}
        & \includegraphics[width=0.15\textwidth]{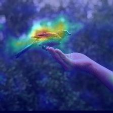}\\

        \rotatebox{90}{{\textbf{RFEM}}}
        & \includegraphics[width=0.15\textwidth]{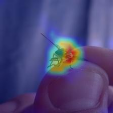} 
        & \includegraphics[width=0.15\textwidth]{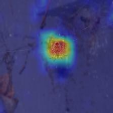} 
        & \includegraphics[width=0.15\textwidth]{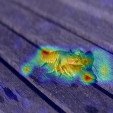}
        & \includegraphics[width=0.15\textwidth]{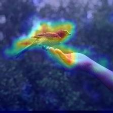}\\

        \rotatebox{90}{\textbf{{RFEM-Class}}}
        & \includegraphics[width=0.15\textwidth]{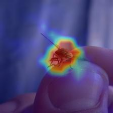} 
        & \includegraphics[width=0.15\textwidth]{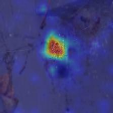} 
        & \includegraphics[width=0.15\textwidth]{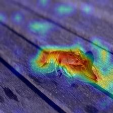}
        & \includegraphics[width=0.15\textwidth]{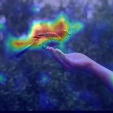}\\
    \end{tabular}
    }
    \caption{Qualitative comparison of different XAI methods for images from the ImageNet1K dataset. Each row corresponds to a specific explanation method.}
    \label{fig:imagenet_visualization}
\end{figure}

\end{document}